\documentclass[runningheads]{llncs}
\usepackage{graphicx}
\usepackage[caption=false]{subfig}
\usepackage[pagebackref=false,breaklinks=true,colorlinks=true,bookmarks=false]{hyperref}
\usepackage{amsmath}
\usepackage{amssymb}
\usepackage{multirow}
\usepackage[misc,geometry]{ifsym}

\begin{document}

\title{Towards Book Cover Design via Layout Graphs}
\titlerunning{Towards Book Cover Design via Layout Graphs}

\author{Wensheng Zhang\inst{1}(\Letter) \and
Yan Zheng\inst{1} \and
Taiga Miyazono\inst{1}\and
Seiichi Uchida\inst{1}\orcidID{0000-0001-8592-7566} \and 
Brian Kenji Iwana\index{Iwana,Brian Kenji}\inst{1}\orcidID{0000-0002-5146-6818}}
\authorrunning{W. Zhang et al.}
\institute{Kyushu University, Fukuoka, Japan
\email{\{zhang.wensheng,yan.zheng,taiga.miyazono\}@human.ait.kyushu-u.ac.jp}\\
\email{\{uchida,iwana\}@ait.kyushu-u.ac.jp}}

\maketitle              
\begin{abstract}
Book covers are intentionally designed and provide an introduction to a book. However, they typically require professional skills to design and produce the cover images. Thus, we propose a generative neural network that can produce book covers based on an easy-to-use layout graph. The layout graph contains objects such as text, natural scene objects, and solid color spaces. This layout graph is embedded using a graph convolutional neural network and then used with a mask proposal generator and a bounding-box generator and filled using an object proposal generator. Next, the objects are compiled into a single image and the entire network is trained using a combination of adversarial training, perceptual training, and reconstruction. Finally, a Style Retention Network (SRNet) is used to transfer the learned font style onto the desired text. Using the proposed method allows for easily controlled and unique book covers.  

\end{abstract}

\section{Introduction} \label{sec:intro}
Book covers are designed to give potential readers clues about the contents of a book. 
As such, they are purposely designed to serve as a form of communication between the author and the reader~\cite{drew2005by}.
Furthermore, there are many aspects of the design of a book cover that is important to the book.
For example, the color of a book cover has shown to be a factor in selecting books by potential readers~\cite{gudinavivcius2018choosing}, the objects and photographs on a book cover are important for the storytelling~\cite{kratz1994telling}, and even the typography conveys information~\cite{tschichold1998new,el2018representing}. 
Book covers~\cite{iwana2016judging,Lucieri_2020} and the objects~\cite{jolly2018how} on book covers also are indicators of genre. 

While book cover design is important, book covers can also be time-consuming to create. 
Thus, there is a need for easy-to-use tools and automated processes which can generate book covers quickly. 
Typically, non-professional methods of designing book covers include software or web-based applications. 
There are many examples of this, such as Canva~\cite{canva}, fotor~\cite{fotor}, Designhill~\cite{designhill}, etc.
These book cover designers either use preset templates or builders where the user selects from a set of fonts and images. 
The issue with these methods is that the design process is very restrictive and new designs are not actually created. It is possible for multiple authors to use the same images and create similar book covers.

Recently, there has been an interest in machine learning-based generation. 
However, there are only a few examples of book cover-based generative models. 
In one example, the website deflamel~\cite{deflamel} generates designs based on automatically selected background and foreground images and title font. The images and font are determined based on a user-entered description of the book plus a ``mood.'' 
The use of Generative Adversarial Networks (GAN)~\cite{goodfellow2014gan} have been used to generate books~\cite{Lucieri_2020,booksby}. 
Although, in the previous GAN-based generation methods, the created book covers were uncontrollable and generate gibberish text and scrambled images.

\begin{figure}[t]
    \centering
    \includegraphics[width=0.95\linewidth]{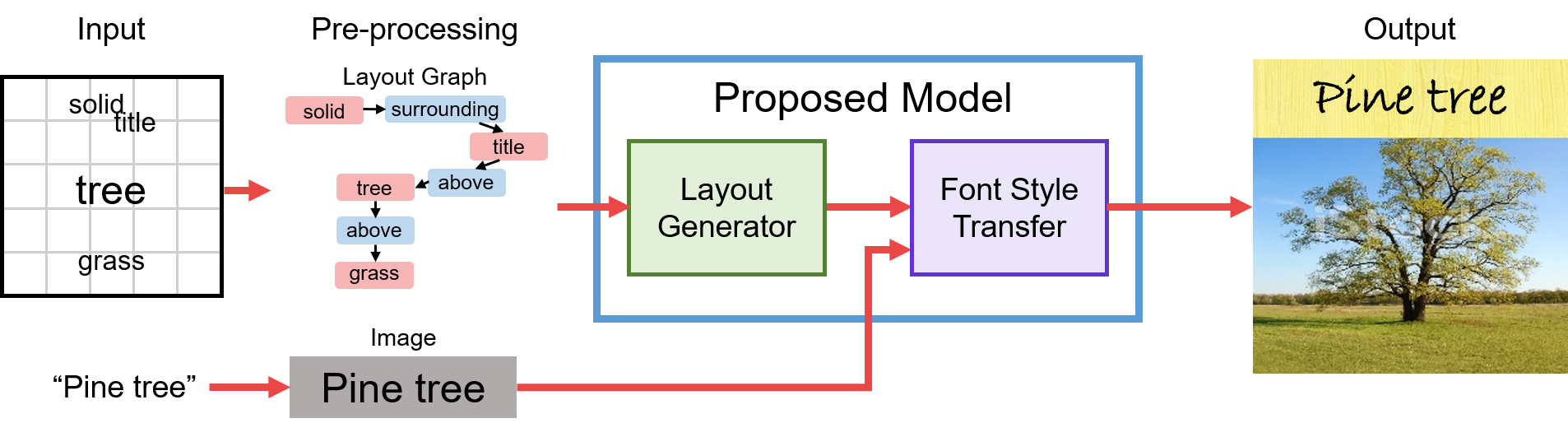}
  \caption{Overview of generating a book cover using a layout graph.}
  \label{fig:layoutgraph}
\end{figure}

The problem with template-based methods is that new designs are not created and the problem with GAN-based methods is that it is difficult to control which objects are used and where they are located. 
Thus, we propose a method to generate book covers that addresses these problems.
In this paper, we propose the use of a \textit{layout graph} as the input for users to draw their desired book cover. 
The layout graph, as shown in Fig.~\ref{fig:layoutgraph}, indicates the size, location, positional relationships, and appearance of desired text, objects, and solid color regions. 
The advantage of using the layout graph is that it is easy to describe a general layout for the proposed method to generate a book cover image from.

\begin{figure}[t]
    \centering
    \includegraphics[width=1.0\linewidth]{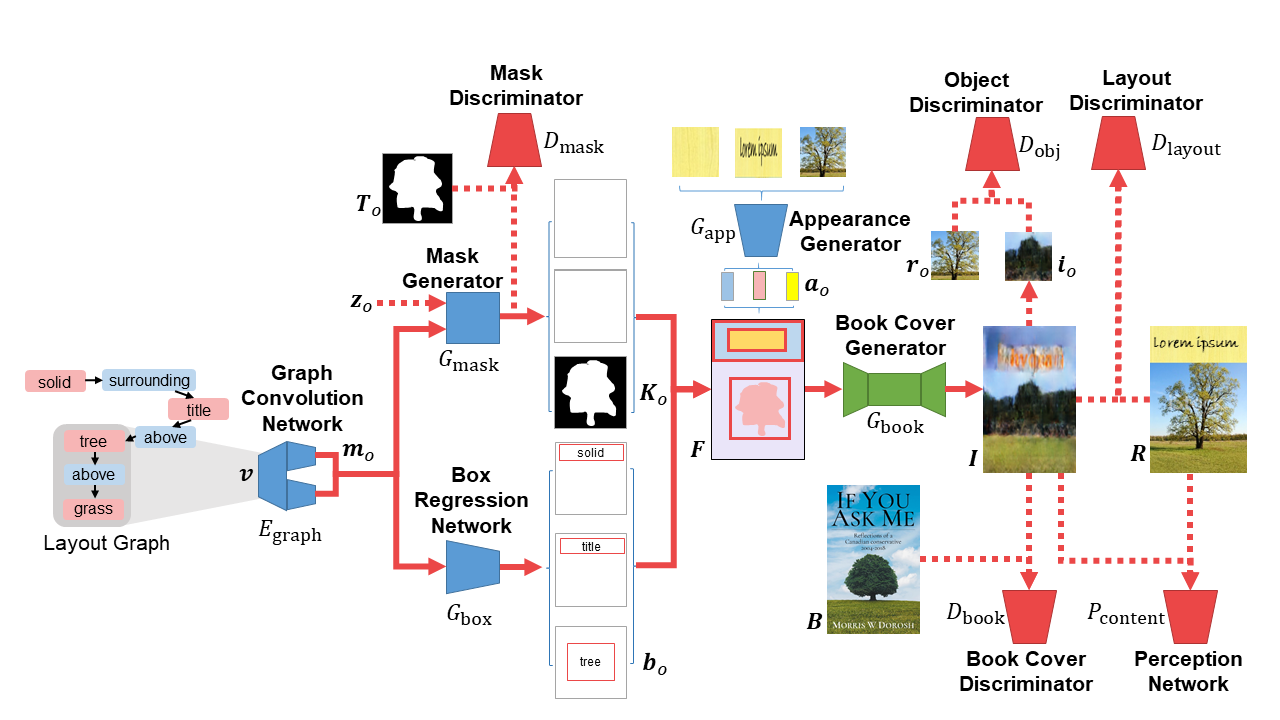}
    \caption{The Layout Generator.}
    \label{fig:full}
\end{figure}

In order to generate the book cover image, the layout graph is provided to a generative network based on scene graph-based scene generators~\cite{Johnson_2018,Ashual_2019}.
In Fig.~\ref{fig:full}, the layout graph is fed to a Graph Convolution Network~(GCN)~\cite{Scarselli_2009} to learn an embedding of the layout objects (i.e. text objects, scene objects, and solid regions). 
This embedding is used to create mask and bounding-box proposals using a mask generator and box regression network, respectively. 
Like~\cite{Ashual_2019}, the mask proposals are used with an appearance generator to fill in the masks with contents. 
The generated objects are then aggregated into a single book cover image using a final generator. 
These generators are trained using four adversarial discriminators, a perception network, and L1 loss to a ground truth image.
Finally, the learned text font is transferred to the desired text using a Style Retention Network~(SRNet)~\cite{Wu_2019}.

The main contributions of this paper are summarized as follows:
\begin{itemize}
    \item As far as the authors know, this is the first instance of using a fully generative model for creating book cover images while being able to control the elements of the cover, such as size, location, and appearance of the text, objects, and solid regions. 
    \item We propose a method of using a combination of a layout graph-based generator and SRNet to create user-designed book cover images.
\end{itemize}
Our codes are shown at \url{https://github.com/Touyuki/Cover_generation}

\section{Related Work}\label{sec:related}
\subsection{Document Generation}
There are many generative models for documents. 
For example, automatic text and font generation is a key task in document generation. 
In the past, models such as using interpolation between multiple fonts~\cite{Campbell_2014,Uchida_2015} and using features from examples~\cite{Suveeranont_2010} have been used. 
More recently, the use of GANs have been used for font generation~\cite{Abe_2017,Hayashi_2019} and neural font style transfer~\cite{Atarsaikhan_2017} has become an especially popular topic in document generation. 

There have also been attempts at creating synthetic documents using GANs~\cite{Bui_2019,Rusticus_2019} and document layout generation using recursive autoencoders~\cite{Patil_2020}.
Also, in a similar task to the proposed method, Hepburn et al. used a GAN to generate music album covers~\cite{hepburn2017album}. 

However, book cover generation, in particular, is a less explored area. 
Lucieri et al.~\cite{Lucieri_2020} generated book covers using a GAN for data augmentation and the website Booksby.ai~\cite{booksby} generated entire books, including the cover, using GANs.
However, while the generated book covers have features of book covers and have the feel of book covers, the objects and text are completely unrecognizable and there is little control over the layout of the cover.

\subsection{Scene Graph Generation}

The proposed layout graph is based on scene graphs for natural scene generation. 
Scene graphs are a subset of knowledge graphs that specifically describe natural scenes, including objects and the relationships between objects. 
They were originally used for image retrieval~\cite{Johnson_2015} but were expanded to scene graph-based generation~\cite{Johnson_2018}. 
In scene graph generation, an image is generated based on the scene graph. 
Since the introduction of scene graph generation, there has been a huge boom of works in the field~\cite{xu2020survey}.
Some examples of scene graph generation with adversarial training, like the proposed method, include Scene Graph GAN (SG-GAN)~\cite{klawonn2018generating}, the scene generator by Ashual et al.~\cite{Ashual_2019}, and PasteGAN~\cite{li2019pastegan}.
These methods combine objects generated by each node of the scene graph and use a discriminator to train the scene image as a whole.
As far as we know, we are the first to propose the use of scene graphs for documents.

\section{Book Cover Generation}

In this work, we generate book covers using a combination of two modules. 
The first is a Layout Generator. 
The Layout Generator takes a \textit{layout graph} and translates it into an initial book cover image. 
Next, the neural font style transfer method, SRNet~\cite{Wu_2019}, is used to edit the generated placeholder text into a desired book cover text or title.

\subsection{Layout Generator}
The purpose of the Layout Generator is to generate a book cover image including natural scene objects, solid regions (margins, headers, etc.), and the title text. 
To do this, we use a layout graph-based generator which is based on scene graph generation~\cite{Johnson_2018,Ashual_2019}. 
As shown in Fig.~\ref{fig:full}, the provided layout graph is given to a comprehensive model of an embedding network, four generators, four discriminators, and a perceptual consistency network. 
The output of the Layout Generator is a book cover image based on the layout graph.

\subsubsection{Layout Graph.}
The input of the Layout Generator is a layout graph, which is a directed graph with each object $o$ represented by a node $\mathbf{n}_o=(\mathbf{c}_o, \mathbf{l}_o)$, where $\mathbf{c}_o$ is a class vector and $l_o$ is the location vector of the object. 
The class vector contains a 128-dimensional embedding of the class of the object.
The location vector $\mathbf{l}_o$ is a 35-dimensional binary vector that includes the location and size of the object.  The first 25 bits of $\mathbf{l}_o$ describe the location of the object on a $5\times5$ grid and the last 10 bits indicate the size of the desired object on a scale of 1 to 10. 

The edges of the layout graph are the positional relations between the objects. 
Each edge $\mathbf{e}_{o,p}$ contains a 128-dimensional embedding of six relationships between every possible pairs of nodes $o$ and $p$. 
The six relationships include ``right of'', ``left of'', ``above'', ``below'', ``surrounding'' and ``inside''. 

\subsubsection{Graph Convolution Network.}
The layout graph is fed to a GCN~\cite{Scarselli_2009}, $E_\mathrm{graph}$, to learn an embedding $\mathbf{m}_o$ of each object $o$. 

Where a traditional Convolutional Neural Network (CNN)~\cite{lecun1998gradient} uses convolutions of shared weights across an image, a GCN's convolutional layers operate on graphs. 
They do this by traversing the graph and using a common operation on the edges of the graph.

\begin{figure}[t]
    \centering
    \subfloat[GCN (Edge)]{
        \includegraphics[width=0.3\columnwidth]{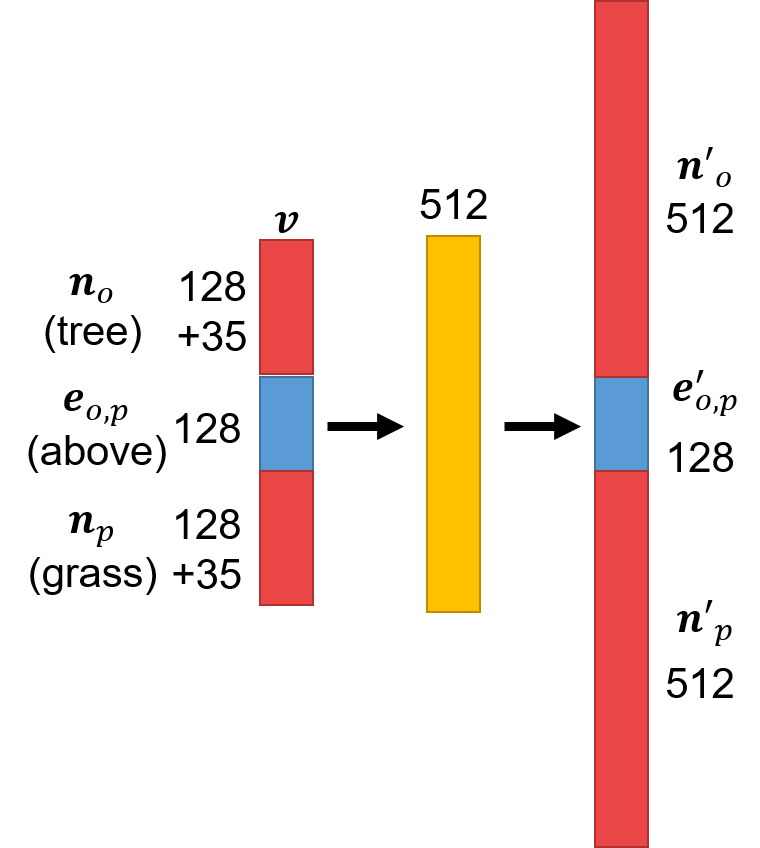}
        \label{fig:GCN-edge}
    }
    \subfloat[GCN (Vertex)]{
        \includegraphics[width=0.3\columnwidth]{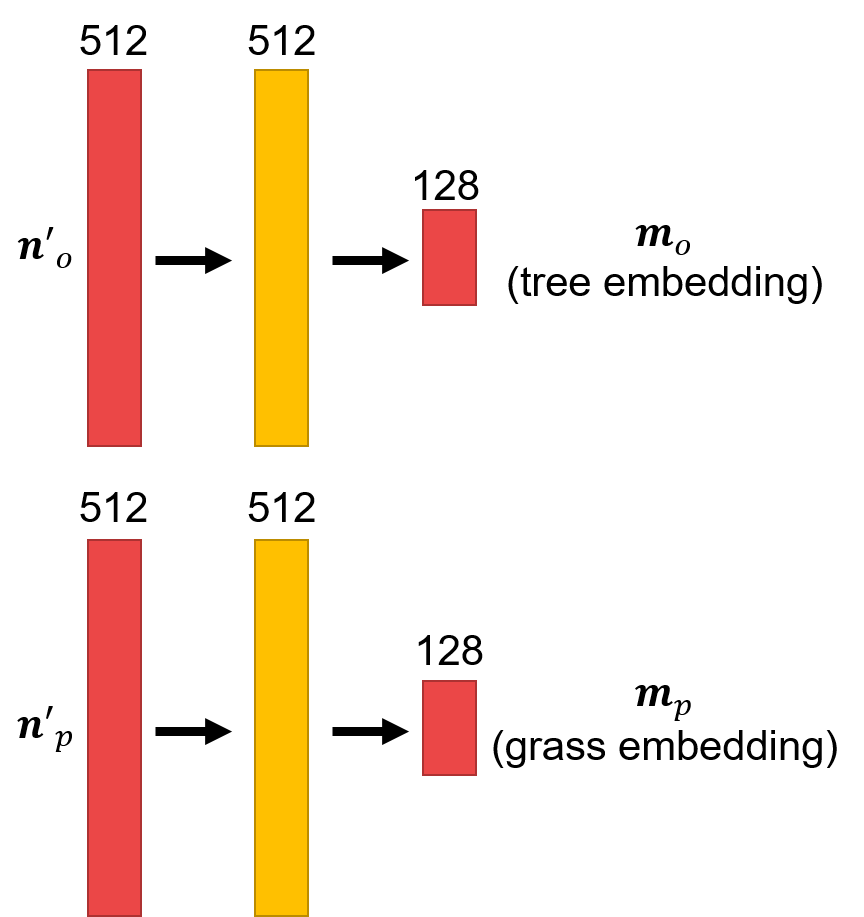}
        \label{fig:GCN-vertex}
    }
    \caption{Illustration of the Graph Convolution Network. The red boxes are vertex vectors, the blue is the edge vector, the yellow is a hidden layer, and the arrows are full connections.}\label{fig:GCN}
\end{figure}

To construct the GCN, we take the same approach as Johnson et al.~\cite{Johnson_2018} which constructs a list of all of the nodes and edges in combined vectors $\mathbf{v}$ and then uses a multi-layer perceptron (MLP) on the vectors, as shown in Fig.~\ref{fig:GCN}. 
Vector $\mathbf{v}$ consists of a concatenation of an edge embedding $\mathbf{e}_{o,p}$ and the two adjacent vertices $o$ and $p$ and vertex embeddings $\mathbf{n}_o$ and $\mathbf{n}_p$. 
The GCN is consists of two sub-networks. 
The GCN (Edge) network in Fig.~\ref{fig:GCN-edge} takes in vector $\mathbf{v}$ and then performs the MLP operation. 
The output is then broken up into temporary object segments $\mathbf{n}'_o$ and $\mathbf{n}'_p$ and further processed by individual GCN (Vertex) networks for each object. 
The result of GCN (Vertex) is a 128-dimensional embedding for each object, which is used by the subsequent Box Regression Network and Mask Generator.

\subsubsection{Mask Generator and Discriminator.}
The purpose of the Mask Generator is to generate a mask of each isolated object for the Appearance Generator. 
The Mask Generator is based on a CNN.
The input of the Mask Generator is the object embedding $\mathbf{m}_o$ learned from the GCN and the output is a $32\times32$ shape mask of the target object. 
This mask is only the shape and does not include size information.
Furthermore, since the Mask Generator creates detailed masks, a variety of shapes should be used. To do this, a 64-dimensional random vector $\mathbf{z}_o$ is concatenated with the object embedding $\mathbf{m}_o$ before being given to the Mask Generator.

In order to produce realistic object masks, an adversarial Mask Discriminator $D_{\mathrm{mask}}$ is used. 
The Mask Discriminator is based on a conditional Least Squares GAN (LS-GAN)~\cite{mao2017least} with the object class $s_o$ as the condition. 
It should be noted that the object class $\mathbf{s}_o$ is different than the 128-dimensional class vector $\mathbf{c}_o$ in the layout graph. 
The GAN loss $\mathcal{L}^D_{mask}$ is: 

\begin{equation}
    \mathcal{L}^D_{\mathrm{mask}}=[\log D_{\mathrm{mask}}(\mathbf{T}_o,\mathbf{s}_o)]+\mathbb{E}_{\mathbf{z}_o\sim\mathcal{N}(0,1)^{64}} [\log(1-D_{\mathrm{mask}}(G_{\mathrm{mask}}(\mathbf{m}_o,\mathbf{z}_o),\mathbf{s}_o)],
\end{equation}
where $G_\mathrm{mask}$ is the Mask Generator and $\mathbf{T}_o$ is a real mask. 
Accordingly, the Mask Discriminator $D_\mathrm{mask}$ is trained to minimize $-\mathcal{L}^D_{mask}$.

\subsubsection{Box Regression Network.}

The Box Regression Network generates a bounding box estimation of where and how big the object should be placed in the layout. 
Just like the Mask Generator, the Box Regression Network receives the object embedding $\mathbf{m}_o$. 
The Box Regression Network is an MLP that predicts the bounding box $\mathbf{b}_o=\{(x_0, y_0), (x_1, y_1)\}$ coordinates for each object $o$.

To generate the layout, the outputs of the Mask Generator and the Box Regression Network are combined. 
In order to accomplish this, the object masks from the Mask Generator are shifted and scaled according to bounding boxes. 
The shifted and scaled object masks are then concatenated in the channel dimension and used with the Appearance Generator to create a layout feature map $F$ for the Book Cover Generator.

\subsubsection{Appearance Generator.}
The objects' appearances that are bound by the mask are provided by the Appearance Generator $G_\mathrm{app}$. 
The Appearance Generator is a CNN that takes real images of cropped objects of ($64\times64\times3)$ resolution and encodes the appearance into a 32-dimension appearance vector. 
The appearance vectors $\mathbf{a}_o$ represent objects within the same class and changing the appearance vectors allows the appearance of the objects in the final generated result to be controlled. This gives the network to provide a variety of different object appearances even with the same layout graph. 
A feature map $\mathbf{F}$ is created by compiling the appearance vectors to fill the masks that were shifted and scaled by the bounding boxes.

\subsubsection{Book Cover Generator.}
The Book Cover Generator $G_\mathrm{book}$ is based on a deep Residual Network (ResNet)~\cite{he2016deep} and it generates the final output. 
The network has three parts. 
The first part is the contracting path made of strided convolutions which encodes the features from the feature map $\mathbf{F}$. 
The second part is a series of 10 residual blocks and the final part is an expanding path with transposed convolutions to upsample the features to the final output image $\mathbf{I}$.

\subsubsection{Perception Network.}
In order to enhance the quality of output of the Book Cover Generator a Perception Network is used. 
The Perception Network $P_\mathrm{content}$ is a pre-trained very deep convolutional network~(VGG)~\cite{simonyan2014very} that is only used to establish a perceptual loss $\mathcal{L}^P_\mathrm{content}$~\cite{johnson2016perceptual}. 
The perceptual loss:
\begin{equation}
    \label{eq:percept}
    \mathcal{L}^P_\mathrm{content}= \sum_{u\in\mathcal{U}} \frac{1}{u}\left|P^{(u)}_\mathrm{content}(I)-P^{(u)}_\mathrm{content}(R)\right|
\end{equation}
is the content consistency between the extracted features of the VGG network $P_\mathrm{content}$ given the generated layout image $I$ and a real layout image $R$. In Eq.~\eqref{eq:percept}, $u$ is a layer in the set of layers $\mathcal{U}$ and $P^{(u)}_\mathrm{content}$ is a feature map from $P_\mathrm{content}$ at layer $u$.

\subsubsection{Layout Discriminator.}
The Layout Discriminator $D_\mathrm{layout}$ is a CNN used to judge whether the layout image $\mathbf{I}$ appears realistic given the layout $\mathbf{F}$. In this way, through the compound adversarial loss $\mathcal{L}_\mathrm{layout}$, the generated layout will be trained to be more indistinguishable from images of real layout images $\mathbf{R}$ and real layout feature maps $\mathbf{Q}$. The loss $\mathcal{L}_\mathrm{layout}$ is defined as:
\begin{align}
\begin{split}
    \mathcal{L}_{\mathrm{layout}}=& \log D_\mathrm{layout}(\mathbf{Q},\mathbf{R})+\log(1- D_\mathrm{layout}(\mathbf{Q},\mathbf{I})) \\
    &+\log(1- D_\mathrm{layout}(\mathbf{F},\mathbf{R}))+\log D_\mathrm{layout}(\mathbf{Q}',\mathbf{R})
\end{split}
\end{align}
where $\mathbf{Q}'$ is a second layout with the bounding box, mask, and appearance attributes taken from a different, incorrect ground truth image with the same objects. 
This is used as a poor match despite having the correct objects.
The aim of the Layout Discriminator is to help the generated image $\mathbf{I}$ with ground truth layout $\mathbf{Q}$ to be indistinguishable from real image $\mathbf{R}$.

\subsubsection{Book Cover Discriminator.}
The Book Cover Discriminator is an additional discriminator that is used to make the generated image look more like a book. 
Unlike the Layout Discriminator, the Book Cover Discriminator only compares the generated image $\mathbf{I}$ to random real book covers $\mathbf{B}$. 
Specifically, an adversarial loss:
\begin{equation}
    \mathcal{L}_{\mathrm{book}}= \log D_\mathrm{book}(\mathbf{B}) + \log(1- D_\mathrm{book}(\mathbf{I})),
\end{equation}
where $D_\mathrm{book}$ is the Book Cover Discriminator, is added to the overall loss.

\subsubsection{Object Discriminator.}
The Object Discriminator $D_\mathrm{obj}$ is another CNN used to make each object images look real. $\mathbf{i}_o$ is an object image cut from the generated image by the generated bounding box and $\mathbf{r}_o$ is a real crop from the ground truth image.
The object loss $\mathcal{L}_\mathrm{obj}$ is:
\begin{equation}
    \mathcal{L}_{\mathrm{obj}}=\sum^O_{o=1} \log D_\mathrm{obj}(\mathbf{r}_o)-\log D_\mathrm{obj}(\mathbf{i}_o).
\end{equation}

\subsubsection{Training.}

The entire Layout Generator with all the aforementioned networks are trained together end-to-end. 
This is done using a total loss:
\begin{align}
\begin{split}
\label{eq:total}
\mathcal{L}_\mathrm{total} =& \lambda_1\mathcal{L}_\mathrm{pixel} + \lambda_2\mathcal{L}_\mathrm{box} + \lambda_3\mathcal{L}^P_\mathrm{content} + \lambda_4\mathcal{L}^D_\mathrm{mask} + \lambda_5\mathcal{L}^D_\mathrm{obj} \\ 
& + \lambda_6\mathcal{L}^D_\mathrm{layout}+ \lambda_7\mathcal{L}^D_\mathrm{book} + \lambda_8\mathcal{L}^P_\mathrm{mask} + \lambda_9\mathcal{L}^P_\mathrm{layout}, 
\end{split}
\end{align}
where each $\lambda$ is a weighting factor for each loss. 
In addition to the previously described losses, Eq.~\eqref{eq:total} contains a pixel loss $\mathcal{L}_\mathrm{pixel}$ and two additional perceptual losses $\mathcal{L}^P_\mathrm{mask}$ and $\mathcal{L}^P_\mathrm{layout}$.
The pixel loss $\mathcal{L}_\mathrm{pixel}$ is the L1 distance between the generated image $\mathbf{I}$ and the ground truth image $\mathbf{R}$. 
The two perceptual losses $\mathcal{L}^P_\mathrm{mask}$ and $\mathcal{L}^P_\mathrm{layout}$ are similar to $\mathcal{L}^P_\mathrm{content}$ (Eq.~\ref{eq:percept}), except instead of a separate network, the feature maps of all of the layers of discriminators $D_\mathrm{mask}$ and $D_\mathrm{layout}$ are used, respectively. 

\subsection{Solid Region Generation}

The original scene object generation is designed to generate objects in natural scenes that seem realistic. However, if we want to use it in book cover generation we should generate more elements that are unique to book covers, such as the solid region and the title information.

\begin{figure}[t]
    \centering
    \includegraphics[width=1\linewidth]{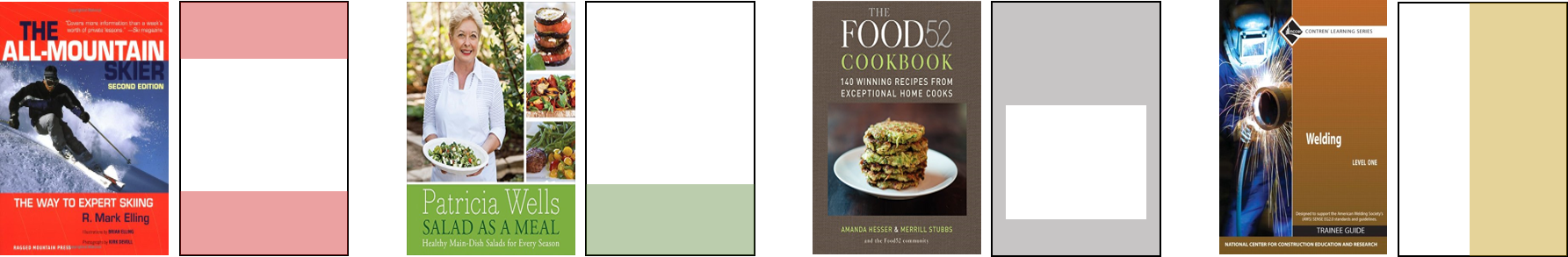}
  \caption{Solid regions.}
  \label{fig:solid}
\end{figure}

We refer to \textit{solid regions} as regions on a book with simple colors. They can be a single solid color, gradients, or subtle designs. As shown in Fig.~\ref{fig:solid}, they are often used for visual saliency, backgrounds, and text regions. 
Except for some text information, usually, there are no other elements in these regions.
To incorporate the solid regions into the proposed model, we prepared solid regions as objects in the Layout Graph. In addition, the solid regions are added as an object class to the various components of the Layout Generator as well as added to the ground truth images $\mathbf{R}$ and layout feature maps $\mathbf{Q}$.
To make sure we can generate realistic solid regions, in our experiment, we used solid regions cut from real book covers.

\subsection{Title Text Generation}
Text information is also an important part of the book covers. It contains titles, sub-titles, author information, and other text. In our experiment, we only consider the title text.

Unlike other objects, like trees, the text information cannot be random variations and has to be determined by the user. 
However, the text still needs to maintain a style and font that is suitable for the book cover image. 

\begin{figure}[t]
    \centering
    \includegraphics[width=0.9\linewidth]{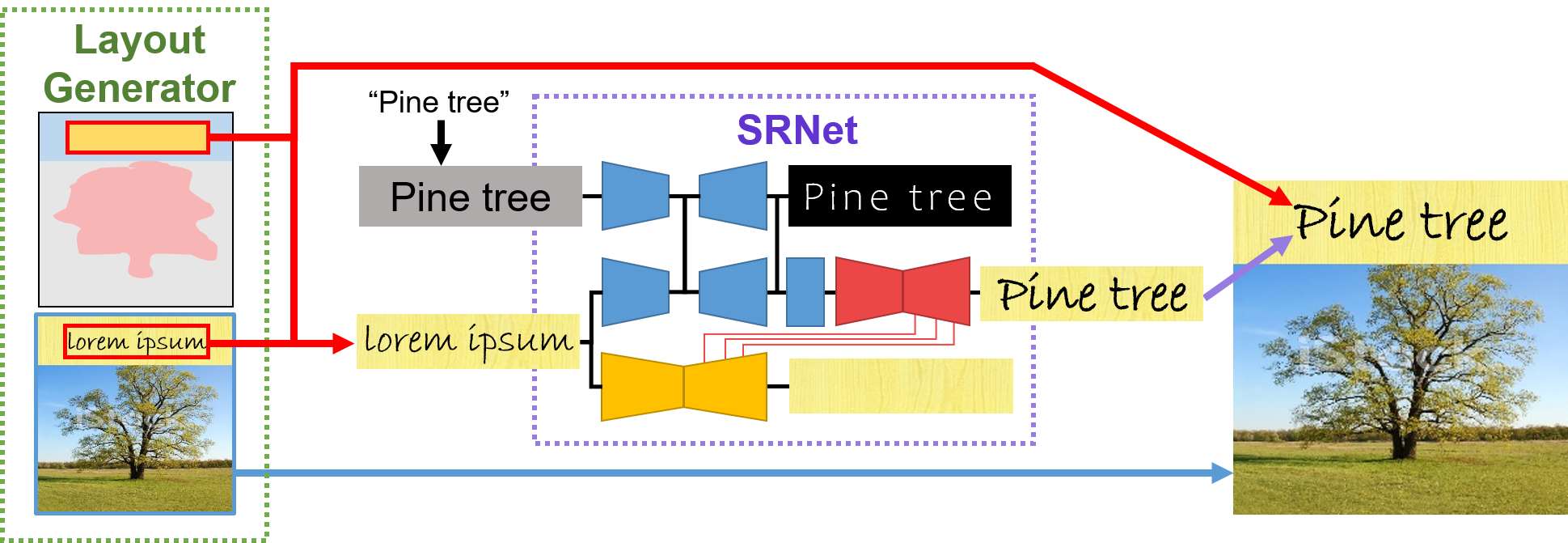}
  \caption{The process of the SRNet.}
  \label{fig:SRnet process}
\end{figure}

Thus, we propose to generate the title text in the image using a placeholder and use font style transfer to transfer the placeholder's font to the desired text. 
Fig.~\ref{fig:SRnet process} shows our process of transferring the font style to the title text. 
To do this, we use SRNet~\cite{Wu_2019}. 
SRNet is a neural style transfer method that uses a skeleton-guided network to transfer the style of text from one image to another. 
In SRNet, there are two inputs, the desired text in a plain font and the stylized text. 
The two texts are fed into a multi-task encoder-decoder that generates a skeleton image and a stylized image of the desired text.
Using SRNet, we can generate any text using the style learned by the Layout Generator and use it to replace the placeholder.

To train the Layout Generator, we use a placeholder text, ``Lorem Ipsum,'' to represent the title. 
Similar to the solid region object, the title object is also added as an object class. 
For the ground truth images $\mathbf{R}$, a random font, color, and location are used. However, the purpose of the Book Cover Discriminator $D_\mathrm{book}$ is to ensure that the combination is realistic as books.

\section{Experimental Results}

\subsection{Dataset}
To train the proposed method two datasets are required. 
The first is the Book Cover Dataset\footnote{\url{https://github.com/uchidalab/book-dataset}}. 
This dataset is made of book cover images and is used to train the Book Cover Discriminator. 

For the second dataset, a natural scene object dataset with semantic segmentation information is required. 
For this, we use 5,000 images from the COCO\footnote{\url{https://cocodataset.org/}} dataset.
For the ground truth images and layouts, random solid regions and titles are added. 
The cropped parts of COCO are used with the Mask Discriminator and the Object Discriminator, and modified
images of COCO are used for the Layout Discriminator and Perception Network. All of the images are resized to $128\times128$.

\subsection{Settings and Architecture}

The networks in the Layout Generator are trained end-to-end using Adam optimizer~\cite{kingma2014adam} with $\beta=0.5$ and an initial learning rate of 0.001 for 100,000 iterations with batch size 6. 
For the losses, we set $\lambda_1, \lambda_4, \lambda_6, \lambda_7=1$, $\lambda_2,\lambda_3,\lambda_8,\lambda_9=10$, and $\lambda_5=0.1$.
The hyperparameters used in the experiments are listed in Table~\ref{tab:architecture}.
For SRNet, we used a pre-trained model\footnote{\url{https://github.com/Niwhskal/SRNet}}.

\begin{table}[p]
\caption{The architecture of the networks.}\label{tab:architecture}
\scriptsize
\centering
\begin{tabular}{|l|l|l|l|}
\hline
\textbf{Network} &  \textbf{Layers} & \textbf{Activation} & \textbf{Norm.}\\
\hline
GCN (Edge) & FC, 512 nodes & ReLU & \\
& FC, 1,152 nodes & ReLU & \\
\hline
GCN (Vertex)& FC, 512 nodes & ReLU & \\
& FC, 128 nodes & ReLU & \\
\hline
Box Regression Network & FC, 512 nodes & ReLU & \\
 & FC, 4 nodes & ReLU & \\
\hline
Mask Generator & Conv. ($3\times3$), 192 filters, stride 1 & ReLU & Batch norm\\
& Conv. ($3\times3$), 192 filters, stride 1 & ReLU & Batch norm.\\
& Conv. ($3\times3$), 192 filters, stride 1 & ReLU & Batch norm.\\
& Conv. ($3\times3$), 192 filters, stride 1 & ReLU & Batch norm.\\
& Conv. ($3\times3$), 192 filters, stride 1 & ReLU & Batch norm.\\
\hline
Appearance Generator & Conv. ($4\times4$), 64 filters, stride 2 & LeakyReLU & Batch norm.\\
& Conv. ($4\times4$), 128 filters, stride 2 & LeakyReLU & Batch norm.\\
& Conv. ($4\times4$), 256 filters, stride 2 & LeakyReLU & Batch norm.\\
& Global Average Pooling & & \\
& FC, 192 nodes & ReLU & \\
& FC, 64 nodes & ReLu & \\
\hline
Book Cover Generator & Conv. ($7\times7$), 64, stride 1 & ReLU & Inst. norm. \\
 & Conv. ($3\times3$), 128 filters, stride 2 & ReLU & Inst. norm. \\
 & Conv. ($3\times3$), 256 filters, stride 2 & ReLU & Inst. norm. \\
 & Conv. ($3\times3$), 512 filters, stride 2 & ReLU & Inst. norm. \\
 & Conv. ($3\times3$), 1,024 filters, stride 2 & ReLU & Inst. norm. \\
\cline{2-2}
\multicolumn{1}{|r|}{($\times10$ residual blocks)} & Conv. ($3\times3$), 1,024 filters, stride 1 & ReLU & Inst. norm. \\ 
 & Conv. ($3\times3$), 1,024 filters, stride 1 & ReLU & Inst. norm. \\
\cline{2-2}
 & T. conv. ($3\times3$), 512 filters, stride 2 & ReLU & Inst. norm. \\
 & T. conv. ($3\times3$), 256 filters, stride 2 & ReLU & Inst. norm. \\
 & T. conv. ($3\times3$), 128 filters, stride 2 & ReLU & Inst. norm. \\
 & T. conv. ($3\times3$), 64 filters, stride 2 & ReLU & Inst. norm. \\
 & Conv. ($7\times7$), 3 filters, stride 1 & Tanh & \\
\hline
Mask Discriminator & Conv. ($3\times3$), 64 filters, stride 2 & LeakyReLU & Inst. norm.\\
& Conv. ($3\times3$), 128 filters, stride 2 & LeakyReLU & Inst. norm.\\
& Conv. ($3\times3$), 256 filters, stride 1 & LeakyReLU & Inst. norm.\\
& Conv. ($3\times3$), 1 filters, stride 1 & LeakyReLU & \\
& Ave. Pooling ($3\times3$), stride 2 & & \\
\hline
Layout Discriminator & Conv. ($4\times4$), 64 filters, stride 2 & LeakyReLU &  \\
 & Conv. ($4\times4$), 128 filters, stride 2 & LeakyReLU & Inst. norm. \\
 & Conv. ($4\times4$), 256 filters, stride 2 & LeakyReLU & Inst. norm. \\
 & Conv. ($4\times4$), 512 filters, stride 2 & LeakyReLU & Inst. norm. \\
 & Conv. ($4\times4$), 1 filter, stride 2 & Linear &  \\
 & Conv. ($4\times4$), 64 filters, stride 2 & LeakyReLU &  \\
 & Conv. ($4\times4$), 128 filters, stride 2 & LeakyReLU & Inst. norm. \\
 & Conv. ($4\times4$), 256 filters, stride 2 & LeakyReLU & Inst. norm. \\
 & Conv. ($4\times4$), 512 filters, stride 2 & LeakyReLU & Inst. norm. \\
 & Conv. ($4\times4$), 1 filter, stride 2 & Linear & \\
& Ave. Pooling ($3\times3$), stride 2 & & \\
\hline
Book Cover Discriminator & Conv. ($4\times4$), 64 filters, stride 2 & LeakyReLU &  \\
 & Conv. ($4\times4$), 128 filters, stride 2 & LeakyReLU & Batch norm. \\
 & Conv. ($4\times4$), 256 filters, stride 2 & LeakyReLU & Batch norm. \\
 & Conv. ($4\times4$), 512 filters, stride 2 & LeakyReLU & Batch norm. \\
 & Conv. ($4\times4$), 512 filters, stride 2 & LeakyReLU & Batch norm. \\
 & Conv. ($4\times4$), 1 filter, stride 2 & Sigmoid & \\
\hline
Object Discriminator & Conv. ($4\times4$), 64 filters, stride 2 & LeakyReLU &  \\
 & Conv. ($4\times4$), 128 filters, stride 2 & LeakyReLU & Batch norm. \\
 & Conv. ($4\times4$), 256 filters, stride 2 & LeakyReLU & Batch norm. \\
& Global Average Pooling & & \\
 & FC, 1024 nodes & Linear &  \\
 & FC, 174 nodes & Linear &  \\
\hline
Perception Network & Pre-trained VGG~\cite{simonyan2014very} & & \\
\hline
Font Style Transfer & Pre-trained SRNet~\cite{Wu_2019} & & \\
\hline
\end{tabular}
\end{table}

\begin{figure}[t]
    \centering
    \subfloat[``Pasture'']{
    \begin{minipage}[b]{0.20\textwidth}
    \includegraphics[width=1\columnwidth]{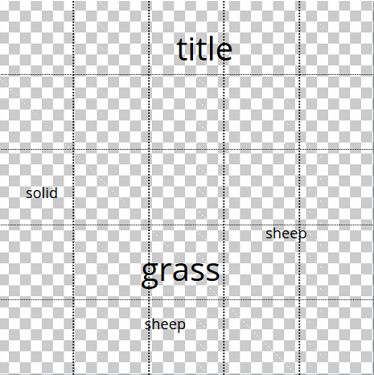}
    \includegraphics[width=1\columnwidth]{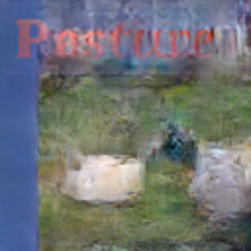}
    \end{minipage}
    }
    \subfloat[``Boat trip'']{
    \begin{minipage}[b]{0.20\textwidth}
    \includegraphics[width=1\columnwidth]{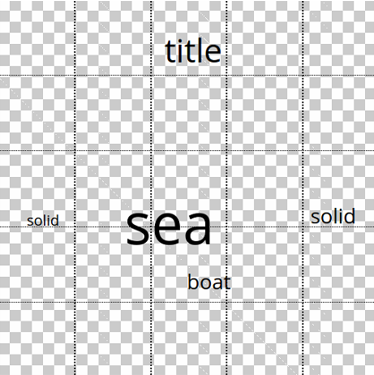}
    \includegraphics[width=1\columnwidth]{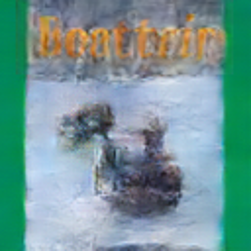}
    \end{minipage}
    }
    \subfloat[``Blue Sky'']{
    \begin{minipage}[b]{0.20\textwidth}
    \includegraphics[width=1\columnwidth]{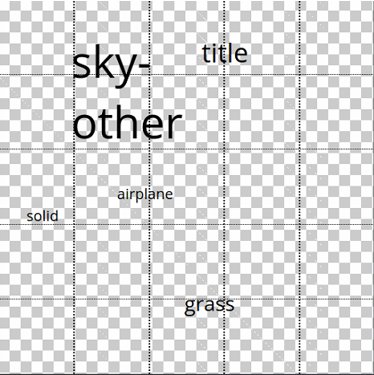}
    \includegraphics[width=1\columnwidth]{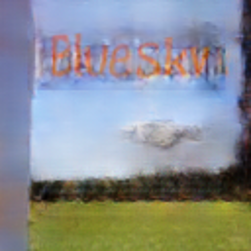}
    \end{minipage}
    }
    \subfloat[``Black Bear'']{
    \begin{minipage}[b]{0.20\textwidth}
    \includegraphics[width=1\columnwidth]{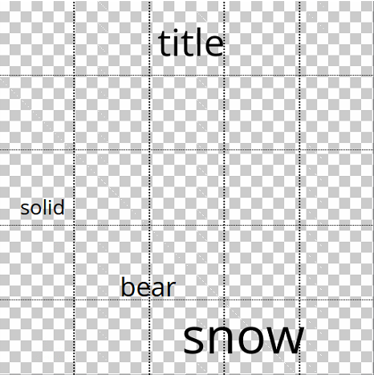}
    \includegraphics[width=1\columnwidth]{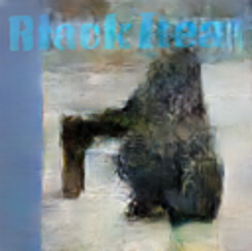}
    \end{minipage}
    }
    
    \subfloat[``Wind'']{
    \begin{minipage}[b]{0.20\textwidth}
    \includegraphics[width=1\columnwidth]{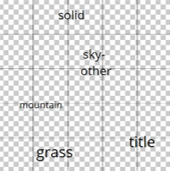}
    \includegraphics[width=1\columnwidth]{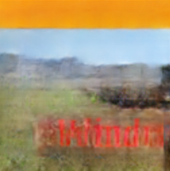}
    \end{minipage}
    }
    \subfloat[``Summer'']{
    \begin{minipage}[b]{0.20\textwidth}
    \includegraphics[width=1\columnwidth]{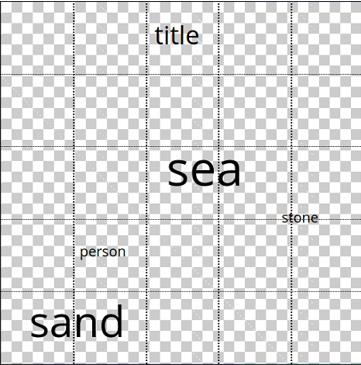}
    \includegraphics[width=1\columnwidth]{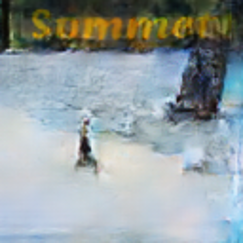}
    \end{minipage}
    }
    \subfloat[``Elephant'']{
    \begin{minipage}[b]{0.20\textwidth}
    \includegraphics[width=1\columnwidth]{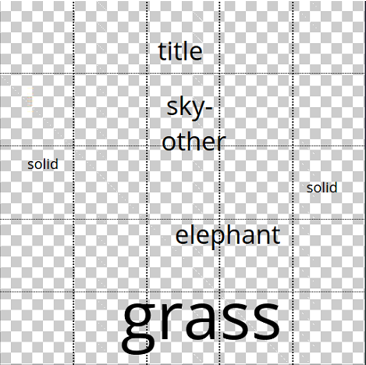}
    \includegraphics[width=1\columnwidth]{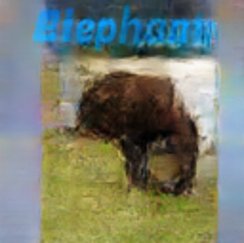}
    \end{minipage}
    }
    \subfloat[``Pizza'']{
    \begin{minipage}[b]{0.20\textwidth}
    \includegraphics[width=1\columnwidth]{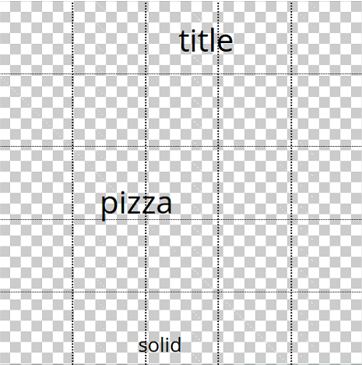}
    \includegraphics[width=1\columnwidth]{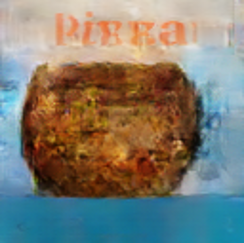}
    \end{minipage}
    }
     \caption{Results with different layouts.}
     \label{fig:Good results}
\end{figure}
\subsection{Generation Results}

\begin{figure}[t]
    \centering
    \subfloat[``Black Bear'']{
    \includegraphics[width=.15\columnwidth]{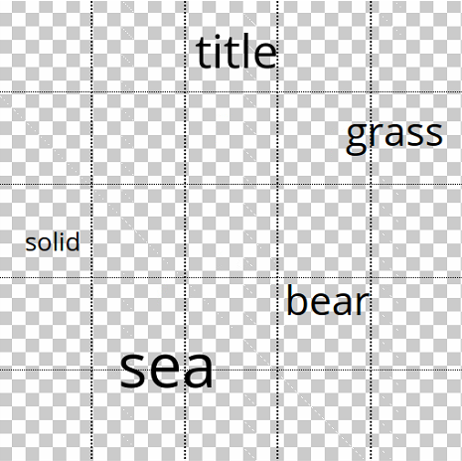} 
    \includegraphics[width=.15\columnwidth]{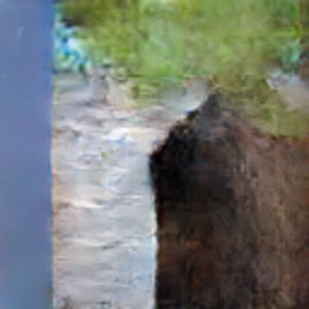}
    }
    \subfloat[``Pasture'']{
    \includegraphics[width=.15\columnwidth]{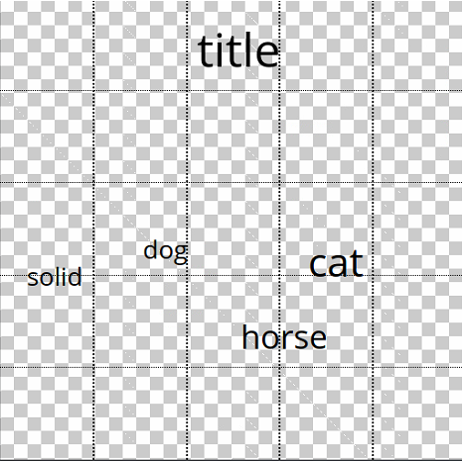}
    \includegraphics[width=.15\columnwidth]{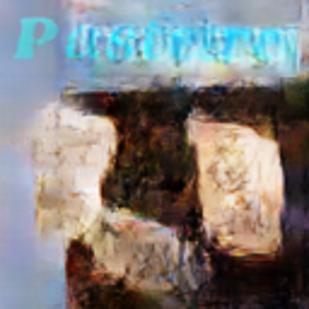}
    }
    \subfloat[``Railway'']{
    \includegraphics[width=.15\columnwidth]{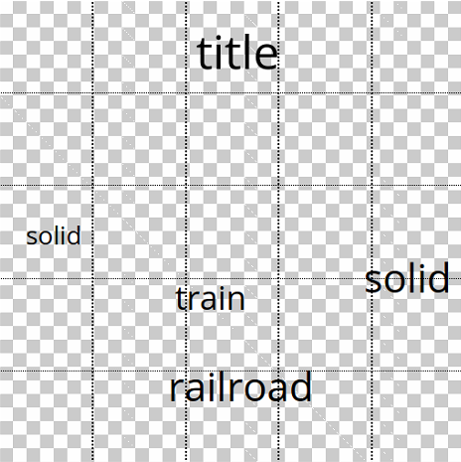}
    \includegraphics[width=.15\columnwidth]{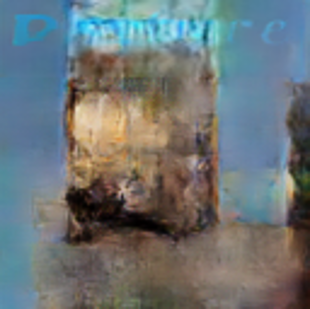}
    }
    \caption{Example of poor results.}
    \label{fig:Bad results}
\end{figure}

Examples of generated book covers are shown in Fig.~\ref{fig:Good results}. 
We can notice that not only the object images can be recognizable, but also the solid regions make the results resemble book covers.
In addition, for most of the results, the generated titles are legible.
While not perfect, these book covers are a big step towards book cover generation.
We also shows some images with poor quality in Fig.~\ref{fig:Bad results}.
In these results the layout maps are reasonable, but the output is still poor. 
This is generally due to having overlapping objects such as ``grass'' on the ``title'' or objects overlapping the solid regions. 

\subsection{Creating Variations in Book Covers}

As mentioned previously, the advantage of using a layout graph is that each node contains information about the object, location, and appearance embedding. 
This allows for the ease of book cover customization using an easy to use interface. 
Thus, we will discuss some of the effects of using the layout graph to make different book cover images.

\subsubsection{Location on the solid region.}
Along with the scene objects, the title text and the solid region can be moved on the layout graph. 
Fig.~\ref{fig:solid} shows examples of generated book covers with the same layout graph except for the ``Solid'' nodes. 
By moving the ``Solid'' node to have different relationship edges with other nodes, the solid regions can be moved. 
In addition, multiple ``Solid'' nodes can be added to the same layout graph to construct multiple solid regions.

\begin{figure}[t]
    \centering
    \includegraphics[width=0.15\columnwidth]{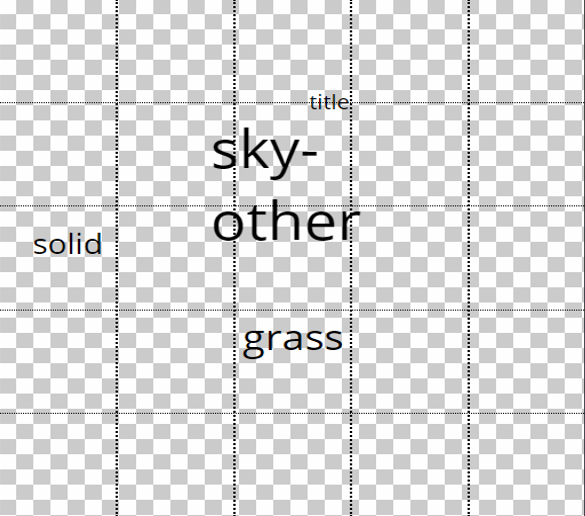}
    \includegraphics[width=0.15\columnwidth]{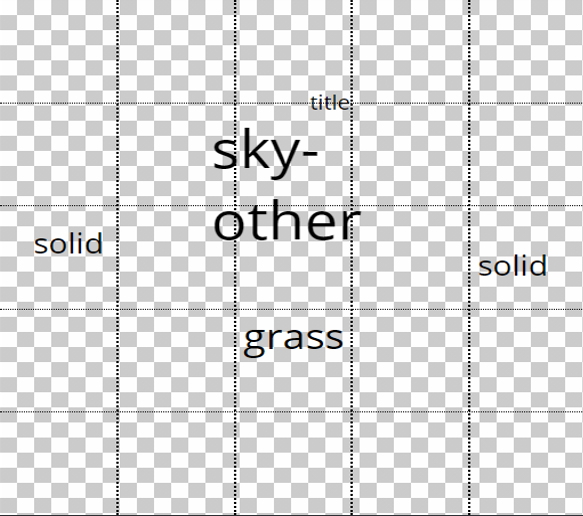}
    \includegraphics[width=0.15\columnwidth]{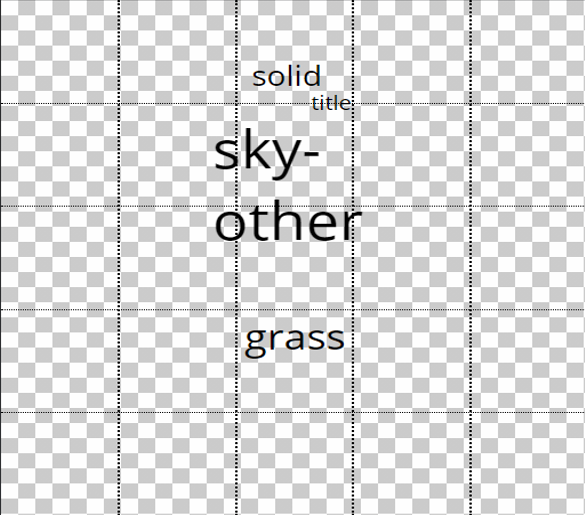}
    \includegraphics[width=0.15\columnwidth]{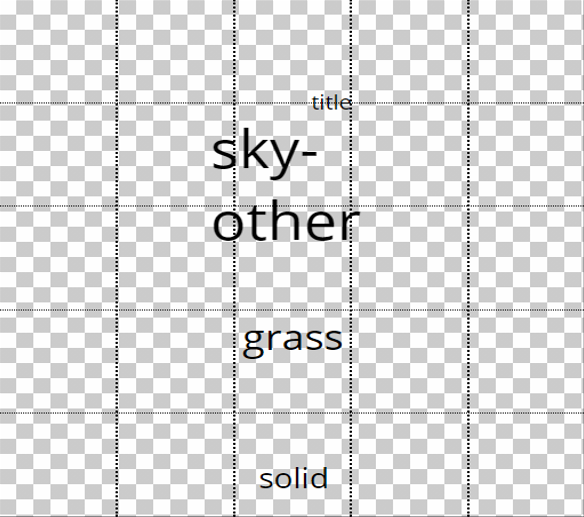}
    \includegraphics[width=0.15\columnwidth]{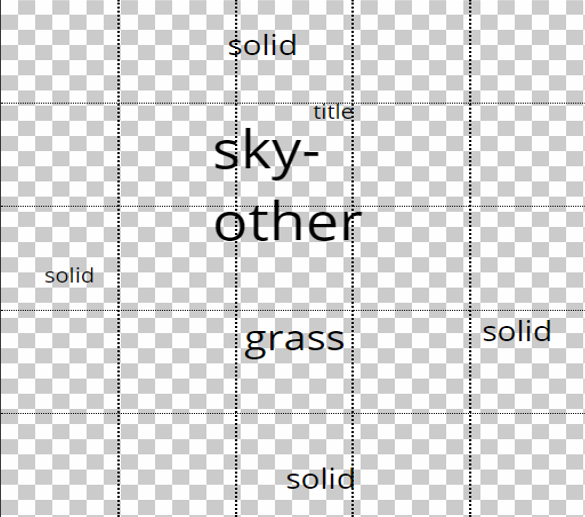}
    
    \includegraphics[width=0.15\columnwidth]{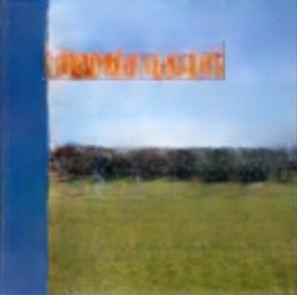}
    \includegraphics[width=0.15\columnwidth]{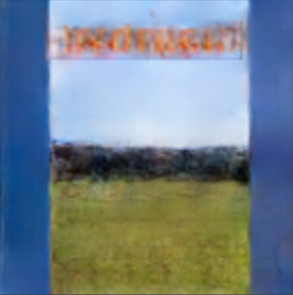}
    \includegraphics[width=0.15\columnwidth]{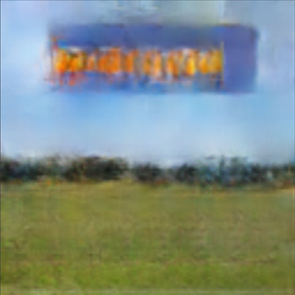}
    \includegraphics[width=0.15\columnwidth]{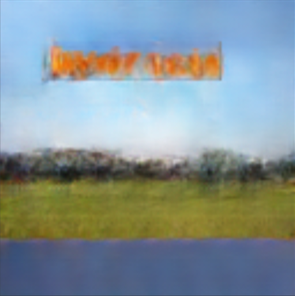}
    \includegraphics[width=0.15\columnwidth]{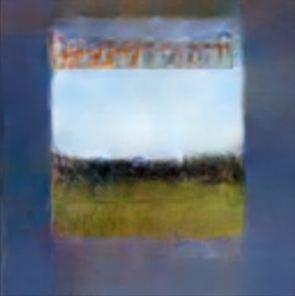}
    \caption{Examples of moving or adding solid region nodes.}\label{fig:solid}
\end{figure}

\subsubsection{Variation in the appearance vector.}
\begin{figure}[t]
    \centering
    \subfloat[Title]{
        \includegraphics[width=0.15\columnwidth]{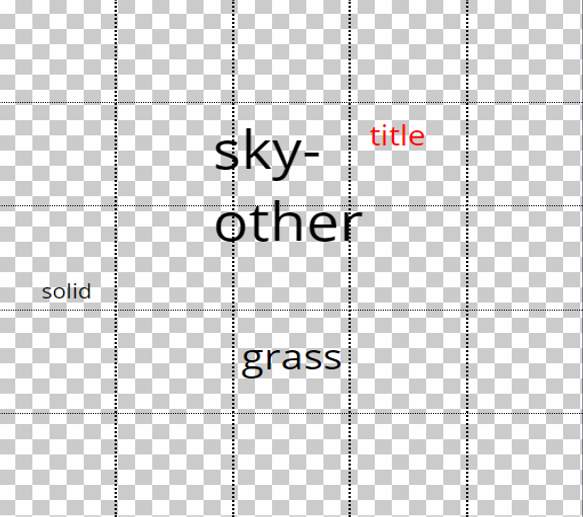}
        \includegraphics[width=0.15\columnwidth]{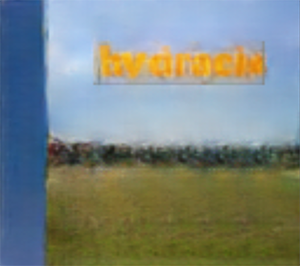}
        \includegraphics[width=0.15\columnwidth]{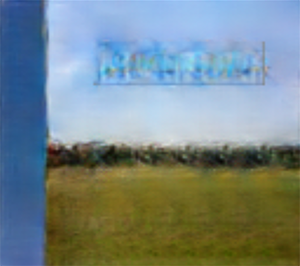}
        \includegraphics[width=0.15\columnwidth]{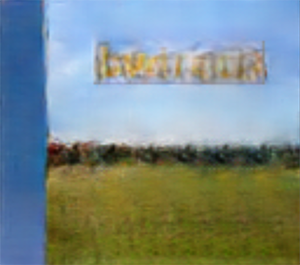}
        \includegraphics[width=0.15\columnwidth]{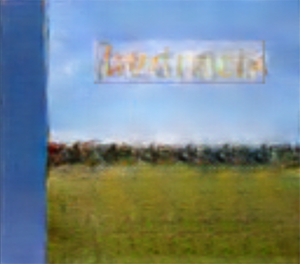}
        \includegraphics[width=0.15\columnwidth]{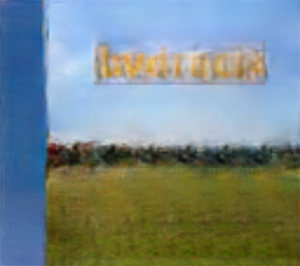}
    }
    
    \subfloat[Solid Region]{
        \includegraphics[width=0.15\columnwidth]{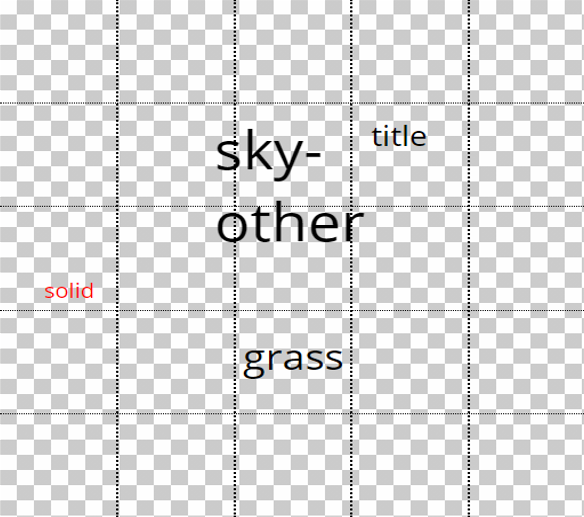}
        \includegraphics[width=0.15\columnwidth]{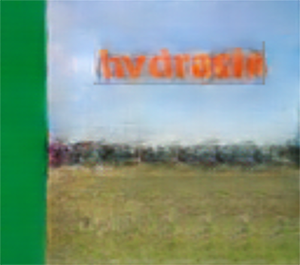}
        \includegraphics[width=0.15\columnwidth]{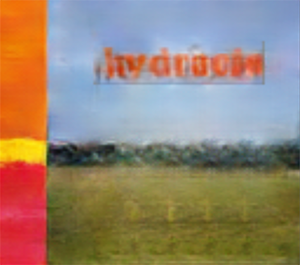}
        \includegraphics[width=0.15\columnwidth]{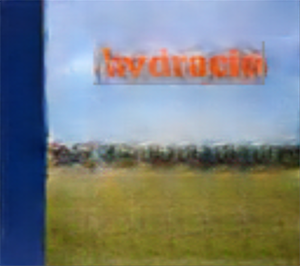}
        \includegraphics[width=0.15\columnwidth]{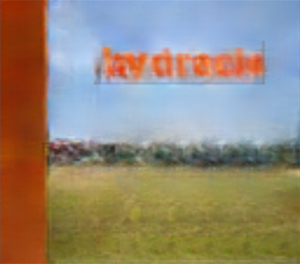}
        \includegraphics[width=0.15\columnwidth]{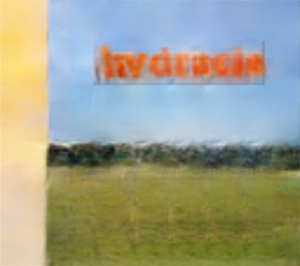}
    }
    
    \subfloat[Grass]{
        \includegraphics[width=0.15\columnwidth]{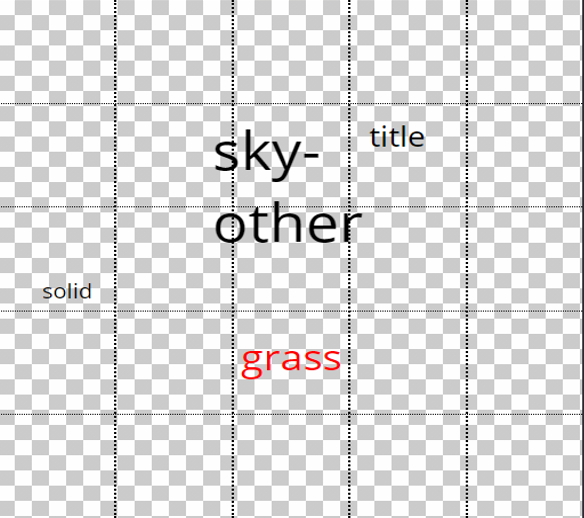}
        \includegraphics[width=0.15\columnwidth]{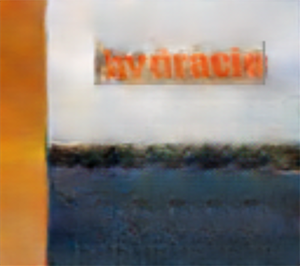}
        \includegraphics[width=0.15\columnwidth]{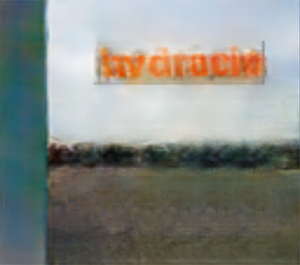}
        \includegraphics[width=0.15\columnwidth]{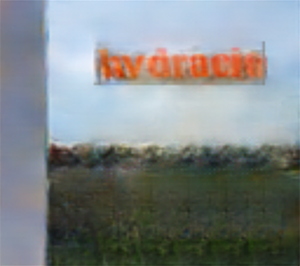}
        \includegraphics[width=0.15\columnwidth]{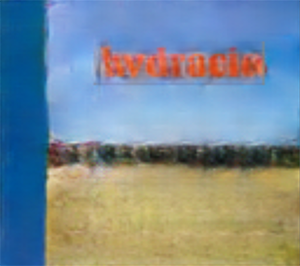}
        \includegraphics[width=0.15\columnwidth]{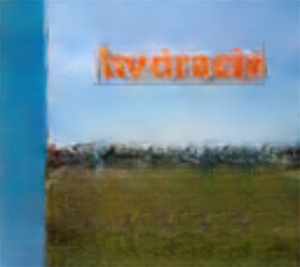}
    }
    
    \subfloat[Sky]{
        \includegraphics[width=0.15\columnwidth]{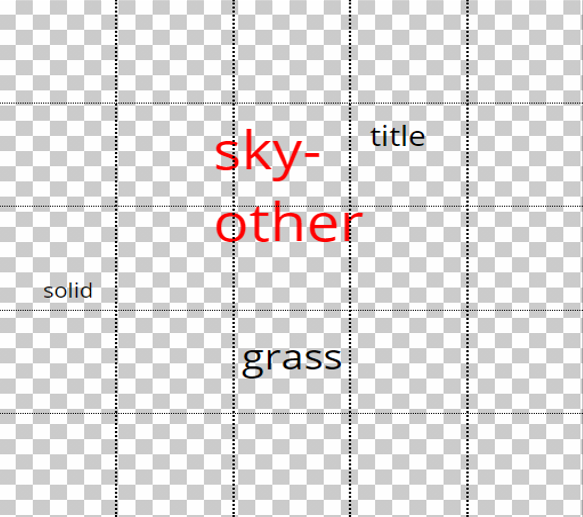}
        \includegraphics[width=0.15\columnwidth]{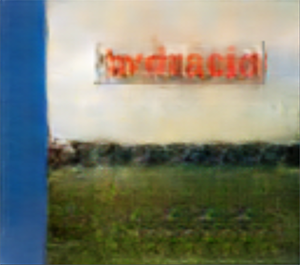}
        \includegraphics[width=0.15\columnwidth]{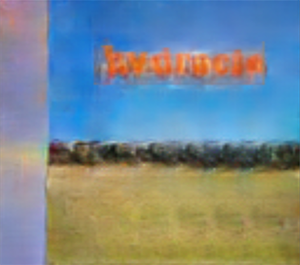}
        \includegraphics[width=0.15\columnwidth]{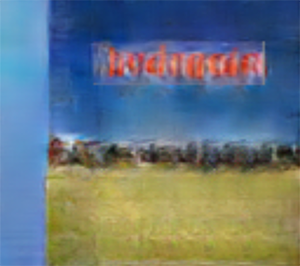}
        \includegraphics[width=0.15\columnwidth]{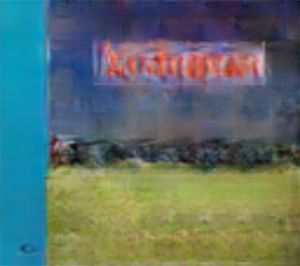}
        \includegraphics[width=0.15\columnwidth]{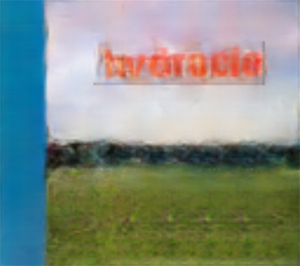}
    }
    \caption{Examples of the effect of changing the appearance vector for different nodes. Each sub-figure changes the appearance vector for the respective node and keeps all other nodes constant.}\label{fig:attribute}
\end{figure}

Due to each node in the layout graph containing its own appearance vector, different variations of generated book covers can be created from the same layout graph. 
Fig.~\ref{fig:attribute} shows a layout graph and the effects of changing the appearance vector of individual nodes. 
In the figure, only one node is changed and all the rest are kept constant.
However, even though only one element is being changed in each sub-figure, multiple elements are affected. 
For example, in Fig.~\ref{fig:attribute}~(c) when changing the ``Grass'' node, the generated grass area changes and the model automatically changes the ``Solid'' and ``Sky'' regions to match the appearance of the ``Grass'' region. 
As it can be observed from the figure, the solid bar on the left is normally contrasted from the sky and the grass.
This happens because each node is not trained in isolation and the discriminators have a global effect on multiple elements and aim to generate more realistic compositions.

\begin{figure}[t]
    \centering
    \includegraphics[width=0.19\columnwidth]{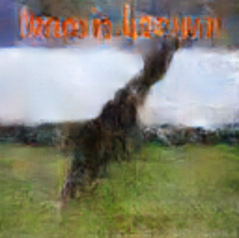}    
    \includegraphics[width=0.19\columnwidth]{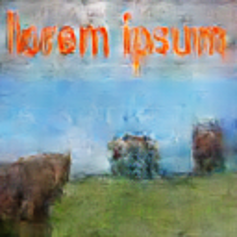}    
    \includegraphics[width=0.19\columnwidth]{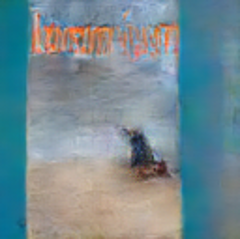}
    \includegraphics[width=0.19\columnwidth]{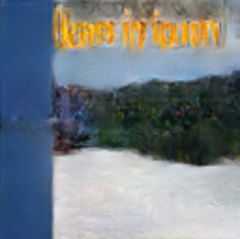}    
    \includegraphics[width=0.19\columnwidth]{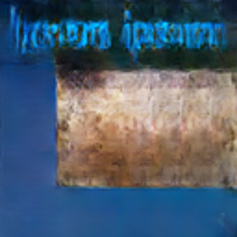}
    
    \includegraphics[width=0.19\columnwidth]{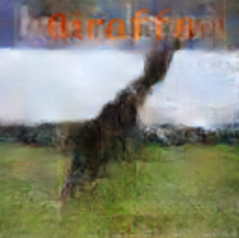}
    \includegraphics[width=0.19\columnwidth]{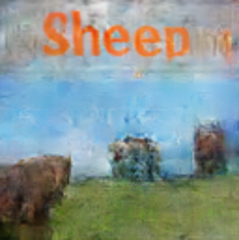}
    \includegraphics[width=0.19\columnwidth]{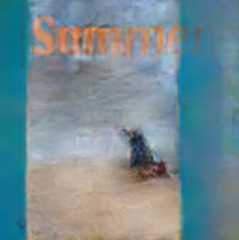}
    \includegraphics[width=0.19\columnwidth]{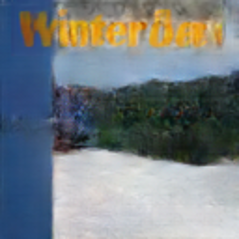}
    \includegraphics[width=0.19\columnwidth]{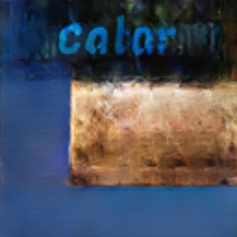}
    
    \setlength{\fboxrule}{0pt}
    \framebox[0.19\columnwidth]{``Giraffe''}
    \framebox[0.19\columnwidth]{``Sheep''}
    \framebox[0.19\columnwidth]{``Summer''}
    \framebox[0.19\columnwidth]{``Winter Day''}
    \framebox[0.19\columnwidth]{``Color''}
    \caption{Using SRNet to change the placeholder title into a target text. The top row is the output before SRNet and the bottom is after SRNet.}\label{fig:comparison}
\end{figure}

\subsection{Effect of Text Style Transfer}
The SRnet is used to change the placeholder text to the desired text in the generated image. In Fig.~\ref{fig:comparison}, we show a comparison of book covers before and after using SRNet. 
As can be seen from the figure, SRNet is able to successfully transfer the font generated by the Layout Generator and apply it to the desired text. 
This includes transferring the color and font features of the placeholder. 
In addition, even if the title text is short like ``Sheep'' or ``Color,'' SRNet was able to still erase the longer placeholder text. 
However, ``Winter Day'' appears to erroneously overlap with the solid region, but that is due to the predicted bounding box of the text overlapping with the solid region. Thus, this is not a result of a problem with SRNet, but with the Box Regression Network.

\section{Conclusion}\label{sec:conclusion}
We proposed a book cover image generation system given a layout graph as the input. It comprises an image generation model and a font style transfer network. The image generation model uses a combination of a GCN, four generators, four discriminators, and a perception network to a layout image. The font style transfer network then transfers the style of the learned font onto a replacement with the desired text. This system allows the user to control the book cover elements and their sizes, locations, and appearances easily. In addition, users can write any text information and fonts fitting the book cover will be generated. Our research is a step closer to automatic book cover generation. 
\section*{Acknowledgement}
This work was in part supported by MEXT-Japan
(Grant No. J17H06100 and Grant No. J21K17808).

\bibliographystyle{splncs04}
\bibliography{icdar}
\end{document}